\documentclass[12pt,reqno]{amsart}

\usepackage[utf8]{inputenc}
\usepackage[T1]{fontenc}
\usepackage[english]{babel}
\usepackage{graphicx}
\usepackage{amsfonts,amsmath,amssymb,amsthm}
\usepackage[colorlinks=true, urlcolor=blue, linkcolor=blue, citecolor=blue, pdftex]{hyperref}
\usepackage{ulem}
\usepackage[a4paper,body={150mm,220mm},centering]{geometry}
\usepackage{soul}

\newtheorem{definition}{Definition}

\newtheorem{remark}[definition]{Remark}

 \usepackage{subcaption}
\usepackage{enumitem}
\begin{document}

\title[Distance Rank Score]{Distance Rank Score: unsupervised filter method for feature selection on imbalanced dataset}

\author[K. Firdová]{Katarína Firdová}
\address{Univ. Grenoble Alpes, Univ. Savoie Mont Blanc, CNRS, LAMA, 73000 Chambéry, France}
\address{Optimistik, 536 rue Costa de Beauregard, 73000 Chambéry, France}
\email{katarina.firdova@optimistik.fr}

\author[C. Labart]{Céline Labart}
\address{Univ. Grenoble Alpes, Univ. Savoie Mont Blanc, CNRS, LAMA, 73000 Chambéry, France}
\email{celine.labart@univ-smb.fr}

\author[A. Martel]{Arthur Martel}
\address{Optimistik, 536 rue Costa de Beauregard, 73000 Chambéry, France}
\email{arthur.martel@optimistik.fr} 

\date{\today}


\maketitle

\begin{abstract}
This paper presents a new filter method for unsupervised feature selection. This method is particularly effective on imbalanced multi-class dataset, as in case of clusters of different anomaly types. Existing methods usually involve the variance of the features, which is not suitable when the different types of observations are not represented equally. Our method, based on Spearman's Rank Correlation between distances on the observations and on feature values, avoids this drawback. The performance of the method is measured on several clustering problems and is compared with existing filter methods suitable for unsupervised data. 
\end{abstract} \hspace{10pt}



\section{Introduction}\label{Introduction}

\subsection{Motivation}\label{Motivation}
Data preprocessing usually requires a step of dimension reduction for various reasons. Dataset can initially contain many dimensions if it comes from a domain where a large number of parameters are measured (e.g. genomics) or if the original dimension of the dataset has been expanded by transformation of the initial parameters in order to obtain new ones, more meaningful for the analysis purpose. It is the case, for example, when performing automatic feature engineering in chronological data, since it can lead to hundreds of features for each time series.

Both cases lead to a situation in which we have to deal with a large multidimensional dataset containing many irrelevant variables. If we want to use the data for further modeling, it may be preferable to reduce the dimension, because models with less input parameters are:

\begin{enumerate}
	\item faster: computations are less complicated as there are less parameters and then less interactions to consider.
	\item easier to explain: the role of an individual input to the output is easier to describe as there is a limited number of inputs.
	\item usually better: model robustness tends to be higher. 
\end{enumerate}

\subsection{Challenges}\label{Challenge}
In this paper we focus on multi-class unsupervised and imbalanced datasets, while the results are also demonstrated on the balanced dataset. We aim to select the most relevant original features keeping information about all present types of observations.  
More precisely, our feature selection method has the following properties:

\subsubsection*{Work on unsupervised data}\label{unsupervised}
Labeled observations are rare in practice, it is therefore important to propose an unsupervised method, i.e. which does not need to be guided by the type labels.

\subsubsection*{Preserve the variability between different types of observations}\label{infoloss}
One of the challenges of the dimension reduction is to lose as little information about original data as possible while keeping as few variables as possible. 
In multi-class data, the objective is to reduce the dimension while preserving information about all present classes, allowing the clustering or other algorithms to correctly separate or analyse the various types of observations.
Besides, the selection should preserve information on the various types of observations even if the size of some groups is very small compared to the size of the whole dataset, e.g. cluster containing rare anomalies.

\subsubsection*{Keep explainable results}\label{explainable}

Common unsupervised dimension reduction techniques like PCA transforms the data into a new coordinate system where most of the variation in the data can be described with fewer dimensions that the initial data.
This approach is not convenient if we aim to clarify the contribution of individual parameters to the output, because we cannot retrospectively explain the contribution of each initial parameter. It is preferable to keep original features. \\

This paper is divided into five parts. Section \ref{related work} lays out the existing work in the field. Section \ref{methodology} presents new method for feature selection. Section \ref{results} compares the results obtained with the new method to those obtained with selected existing methods. The experimental setup is discussed and performance is demonstrated on both balanced and imbalanced multi-class datasets.
Lastly, section \ref{conclusion} concludes the results and provides the final remarks.

\section{Related work }\label{related work}

As described in \cite{kohaviwrapper}, feature selection methods can be divided in two groups: filter methods and wrapper methods. \\

Filter methods evaluate the features importance in the preprocessing step, i.e. independently of the model algorithm. They use various statistical tests and techniques to return a score for each feature. One of the most typical filter method is Maximum Variance. The variance of each feature is computed and features whose variance is below a chosen threshold are removed as they are considered to be of little importance for the model.\\

Wrapper methods use a subset of features to train the model. They evaluate the features subset impact to the model performance and add or remove as many features as needed to improve the results. Forward selection is a greedy wrapper method that starts with an empty model and add one variable with the best single improvement at each step.\\

More recent literature mentions embedded methods which combine filter and wrapper methods. Feature selection method is integrated as a part of the learning algorithm. \\

Recently, several feature selection algorithms have been designed for imbalanced datasets but many of them are supervised (we refer for example to \cite{relief_imbal},  \cite{fs_autoencoders}, \cite{Mindex}, \cite{FS_Kamal2010}, \cite{FS_Cuaya}, \cite{FS_Yin2013}).  \\
In \cite{relief_imbal}, the author extends the RELIEFF algorithm (see \cite{relief}) to the case of imbalanced datasets. In \cite{fs_autoencoders}, the authors use autoencoders ensemble trained only on the majority class to reconstruct all classes. From the analysis of the reconstruction error, features with different values distribution on the minority class are selected. \cite{FS_weightedKNN} proposes a feature reduction method for imbalanced data classification using similarity-based feature clustering with adaptive weighted K-nearest neighbors.
\cite{FS_roughset} proposes a feature selection for imbalanced datasets based on neighborhood rough sets.
\cite{FS_fuzzy}  proposes a method based on Weighted Mutual Information. \\

Papers proposing an unsupervised method for imbalanced datasets are quite rare (see \cite{FS_He2010} and \cite{FS_Alibeigi}). \cite{FS_He2010} is an unsupervised method for two-class datasets which requires to know the proportion of observations in the minor class and to fix the number of relevant features we want to keep. \cite{FS_Alibeigi} proposes an unsupervised method which consists on removing redundant features. To do so, the authors approximate the probability density function (PDF) of each feature and then remove the features having a PDF with a high covering area. \\

Correlation-based approaches for feature selection have been studied in \cite{thesis99}, \cite{thesis21}, \cite{FS_YeJi2016}, \cite{fs_corr_unsupervised}, \cite{Pattanshetti2018UnsupervisedFS} and \cite{FS_KrigingCor}, but without any special interest in imbalanced classes. \\

Considering practical needs and objectives mentioned in the previous section, we focus on unsupervised filter methods which select the most relevant features. We present in detail below two unsupervised filter methods close to our method.

\subsection{Laplacian score}\label{laplacian}
\cite{laplacian} proposes an unsupervised approach for feature selection, called {\it Laplacian score}. The method assigns a score to each feature according to its importance to keep the local structure. The main idea relies on the fact that a relevant feature should have very similar values on similar observations, i.e. those close to each other in the undirected graph. Following this reasoning, a relevant feature $r$ should minimize the Laplacian score

\begin{equation}\label{eq_laplacian}
 \mbox{Laplacian}_r = \frac{ \sum^{n}_{i,j=1} (M_{i,r}-M_{j,r})^2 S_{i,j}  } {V(M_{:,r})},
\end{equation}

where $M$ is a data matrix of size $n \times p$, $n$ is the number of observations, $p$ is the number of features, $M_{i,r}$ represents the value of the $r$-th feature of the $i$-th observation. $S$ is a similarity matrix and $V(M_{:,r})$ is the variance of $r-$th feature. The role of $V(M_{:,r})$ is to prioritize features with large variance as they have a more representative power. \\

The computed score of a feature aims to reflect its locality preserving power because it is assumed that the local structure of the data space is more important for problems such as classification than its global structure. 
Similarity matrix in \cite{laplacian} considers only $k$ nearest neighbors, i.e. $S_{i,j} = 0$ if observation $i$ is not among $k$ nearest neighbors of observation $j$, in order to focus on the local structure. Usually $k$ is chosen to be $5$,  i.e. the five nearest neighbors are taken into account. \\

\subsection{Compactness score}\label{compact_score}

Similarly, \cite{compactness_score} proposes an unsupervised filter method for feature selection based on the idea of keeping the internal structure of the data space.  According to the authors, relevant features should have compact internal structures, i.e. minimize distances within its nearest neighbors. Variance is used to indicate the level of divergence of the samples and select those with relatively large variance as they are more representative. Based on this argumentation, the Compactness score for the feature $r$ can be written as
\begin{equation} \label{eq_cs}
 \mbox{Compactness score}_{r} = \frac{ \sum^{n}_{i=1}   \sum_{j \in S} | M_{i,r} - M_{j,r}|}  {V(M_{:,r}) },
\end{equation}
where $S$ is the set of $k$-nearest neighbors of $M_{i,r}$, the other notations are kept unchanged. The most relevant features should have the lowest Compactness score.

\section{Methodology}\label{methodology}
Laplacian and Compactness scores described in subsections \ref{laplacian} and \ref{compact_score} meet certain requirements on the suitable method of feature selection - they are unsupervised filter methods based on a comprehensive theory. However, they may lead to information loss as the relations between distant observations are not considered. \\

Another disadvantage is the use of the variance as an indicator of a representative feature. Let us consider the case of a dataset with two imbalanced classes and a feature which has similar values on the big group, similar values on the small one, but differentiated values from one group to the other. This feature is important to distinguish the small group, but it has a small total variance. In case of detection of groups of anomalous events (which are usually rare) this can be an important drawback. \\

Ideally, we should consider a feature as relevant if besides its similar behaviour for the close observations it behaves significantly different when the observations are distant. This is the motivation to propose a method which adhere to the mentioned assumption. \\

\subsection{Distance Rank Score}\label{rho}

In order to include the behavior of all observations, we suggest to measure the relation between pairwise total distances and pairwise distances on a specific feature. This relation is measured by computing the correlation between the ranks of the above mentioned distances. Considering the rank of the distances rather than the distances themselves leads to a robust indicator to extreme values. This type of values often arise in case of anomaly detection.\\

Let us keep the same notations as in subsection \ref{laplacian}. In the following, $N$ represents the number of pairs of observations, i.e. $N= \frac {n(n-1)}{2}$, $(V_k^D )_{k\in[1,N]}$ is a vector of total distances and $(V_r^D )_{k\in[1,N]}$ is a vector of the distances on the $r$-th feature.
Formally, we have $V_k^D = \sum^{p}_{r=1} (M_{i,r} - M_{j,r})^2 $ and $V_k^r = (M_{i,r} - M_{j,r})^2,$ where $(i,j)$ is such that $k=n  \times i + j.$ Then,  we compute Spearman's Rank Correlation Coefficient  between $R_{V^D}$ and $R_{V^r}$, which are respectively the rank vectors of $V^D$ and $V^r$. This can be written as

\begin{equation}\label{eq_rho}
\mbox{Distance Rank Score}_r = 1 - \frac{6 } { N ( N^2 - 1) } \sum^{N}_{k=1}  d^2_k
\end{equation}

where $d^2_k$ is the distance between the $k$-th elements of $R_{V^D}$ and $R_{V^r}$.

The Spearman's Rank Correlation can be used for non-linear relationships. Like other correlation coefficients, values vary between $-1$ and $1$. A strong positive correlation, i.e. a Distance Rank Score close to $1$, suggests a high importance of the considered feature.

\section{Experimental results}\label{results}

In this section we demonstrate the feature selection process on several datasets in the case of clustering problems. Although a feature selection as a preprocessing step is not limited only to clustering problems, it is a common approach to illustrate the efficiency of unsupervised feature selection methods as we easily evaluate the quality of clustering results before and after the feature selection step.

We focus on the problem of imbalanced datasets, i.e. datasets where classes contain significantly different numbers of observations. These types of datasets arise in the context of multi-class anomaly detection. In this case, the objective is twofold: detect the anomalies, i.e. separate the clusters of anomalies from the normal one, and distinguish the different types of anomalies, i.e. separate the different clusters of anomalies. 

To show the efficiency of our feature selection method on imbalanced dataset, we consider two multi-class types of problems: time series anomaly detection and image anomaly detection. To test the performance of our method on balanced dataset, we consider a clustering problem in the case of groups of the same size. As we want to compare unsupervised methods we use the label information only for the purpose of the multi-class evaluation.

\subsection{Datasets}\label{dataset_description} 

\begin{table}[!ht]
  \caption{Data used for demonstration in the following subsections, number of observations in the dataset ($n$), number of dimensions ($p$) after preprocessing (see subsection \ref{data_preparation}) and number of classes ($k$). }
 \label{table_data}
\begin{center}
\begin{tabular}{|c||c|c|c|}
  \hline
 Name & n & p & k \\ 
 \hline 
Dataset 1 & 500 & 1043 & 10 \\ 
 \hline
Dataset 2  & 520 & 1002 & 5\\ 
 \hline
MNIST5 & 1785 & 563 & 5 \\
\hline
orlraws10P & 100 & 2740 & 10 \\
\hline
\end{tabular}
\end{center}
\end{table}

 {\it Dataset 1} contains the characteristics of ten different kinds of two-dimen\-sional time series. A large sample of periodical series is completed by few series with unusual behavior such as constant values or white noise in one or both dimensions. This type of anomalies arise for example in the case of continuous industrial processes.
 More details about original series can be found in \cite{explainable_ad}.
 {\it Dataset 1} gathers hundreds of characteristics for each series, extracted by using the tsfresh \cite{tsfresh} Python package. \\
 
 {\it Dataset 2} contains the characteristics of five different kinds of two-dimensional time series. The most numerous group contains time series generated by an autoregressive process with lag 4. They are completed by a few series representing different processes in both dimensions.  As {\it Dataset 1}, such type of data arise in the case of continuous industrial processes. More details can be found in \cite{explainable_ad}.
 {\it Dataset 2} gathers hundreds of characteristics for each series, extracted using the tsfresh Python package \cite{tsfresh}. \\

\begin{remark} 
{\it Dataset 1} and {\it Dataset 2} illustrate the utility of the feature selection process. Original datasets are composed of $2$-dimensional chronological series. One of the approaches to analyse time series is to model their characteristics instead of all original values. However, when we do not know the behavior of the series, it is challenging to find a procedure which automatically extracts only the most relevant characteristics. Tools like tsfresh \cite{tsfresh} automatically compute hundreds of characteristics from each parameter of a multidimensional time series and the role of the feature selection algorithm is to determine the most important features. Consequently, we can proceed with modeling with a reasonable number of characteristics.
\end{remark} 

 {\it MNIST} is a database of greyscale images of handwritten digits which is widely used to evaluate machine learning algorithms. Each digit image is normalized to fit into a $28\times28$ pixel format and is represented by values from 0 to 255 for each pixel.
 {\it MNIST5} is a subset of the original dataset, containing only observations among five digits (from $0$ to $4$). To obtain an imbalanced dataset, we modify {\it MNIST5} to have one major and many minor classes of digits ($0$ is the major class). Minor classes contain only $10$ samples. The major class is unchanged. This modified dataset is named {\it MNIST5m}. \\

 {\it Orlraws10P} is a high dimensional dataset from scikit-feature repository in Python. It contains 10 groups with the same number of observations.

\subsection{Data preparation}\label{data_preparation} 
Features with identical values on all observations are useless in clustering algorithm as they do not carry any useful information which can contribute to separate the clusters correctly. Besides, their zero variance may cause problems in denominator of Equations \eqref{eq_laplacian} and \eqref{eq_cs}. Constant features are therefore removed from the datasets.   \\

Two or more highly correlated features do not improve the ability of clustering. In order to avoid redundant features in the final selection we remove features with standard Pearson correlation coefficient bigger than $0.95$ in absolute value.  \\

As the demonstrated methods use Euclidean metric to compute distances between observations, values should be on the same scale. We use Python MinMaxScaler to transform each feature to a range of $[0,1]$ to avoid issues with varying magnitudes. The choice of MinMaxScaler is motivated by its high sensitivity to the anomalies compared to standardisation techniques.

\subsection{Evaluation}\label{evaluation} 
 \subsubsection{Metrics}\label{metrics}
The performance of the clustering methods is measured by using three well-known metrics (Accuracy, Normalized Mutual Information and $F_1$ measure) which deal with multi-class clustering. Accuracy and Normalized Mutual Information are suitable metrics for balanced datasets whereas $F_1$ measure is a suitable metric for imbalanced datasets.\\

Let $\textbf{CT}$ denote the vector of the true classes and  $\textbf{CP}$ denote the vector of predicted ones, both of length $n$. Note that predicted classes are adjusted to correspond the true classes by using a permutation mapping function.\\

\begin{itemize}
\item Accuracy is defined by
\begin{equation}\label{accuracy}
ACC = \frac{ \sum^{n}_{i=1} \delta (CT_i, CP_i) } {n},
\end{equation}
where $\delta (x,y) = 1$ if $x = y$ and 0 otherwise.  \\

\item Mutual Information is defined by
\begin{equation} \label{nmi}
MI =  \sum_{CT_i \in \textbf{CT}}  \sum_{CP_i \in \textbf{CP}}  p(CT_i , CP_i) log \frac{ p(CT_i , CP_i) } {p(CT_i) p(CP_i)}, 
\end{equation}
where $p(CT_i)$ (resp. $p(CP_i)$) denotes the probability that a randomly chosen observation belongs to the cluster $CT_i$ (resp. to the cluster $CP_i$) and $p(CT_i , CP_i)$ denotes the joint probability that a randomly chosen observation belongs to both clusters. We use Normalized Mutual Information (NMI) which scales MI to values between 0 and 1. \\

\item $F_1$ measure is defined by
\begin{equation} \label{fbeta}
F_1= \frac{ \mbox{Precision} \times \mbox{Recall} } { \mbox{Precision} + \mbox{Recall}}
\end{equation}
where 
\begin{equation} \label{precision}
\mbox{Precision}= \frac{ \mbox{True Positive} } { \mbox{True Positive} + \mbox{False Positive}}
\end{equation}
and
\begin{equation} \label{recall}
\mbox{Recall} = \frac{ \mbox{True Positive} } { \mbox{True Positive} +  \mbox{False Negative}}
\end{equation}\\

In a multiclass case, we compute a $F_1$ score for each class and then these scores are combined to give an overall score. More precisely, the overall score is a weighted-$F_1$ score which is a weighted average of all per-class scores. The score of each class is weighted by the number of samples in this class.

Let $(C_1,\cdots,C_M)$ denote the $M$ clusters, $N_i$ the number of samples of cluster $C_i$, $N$ the total number of observations and $F_1(C_i)$ the $F_1$ score of cluster $C_i$. The weighted $F_1$ score is given by 
\begin{align}\label{eq_FW}
  F^W_1=\sum_{i=1}^M \frac{N_i}{N}F_1(C_i)
\end{align}

\end{itemize}

 \subsubsection{Clustering evaluation}\label{clustering_evaluation}
As mentioned in section \ref{dataset_description}, we consider some datasets containing imbalanced classes, more precisely one major and several minor classes. Misclassification within minor classes can be masked if the observations from the major group are clustered correctly. A choice of a proper evaluation metric avoids this obstacle but its score does not give us more details. In the context of multi-class anomaly detection, it may be interesting to know if a misclassification occurs in the major group or within the minor groups.\\
We then propose two types of clustering evaluation in case of imbalanced datasets:
\begin{itemize}
  \item A global evaluation on the whole dataset using a weighted $F_1$-score (see \eqref{eq_FW})
\item A two step evaluation which is particularly relevant in case of anomaly detection. This two step evaluation consists in: 
\begin{enumerate}
\item merging all minor clusters and evaluating the clustering performance the major cluster vs all other clusters
\item considering only minor clusters and evaluating the clustering performance among them
\end{enumerate}   
\end{itemize}

\begin{remark}
 The two step evaluation approach allows us to evaluate the capacity to separate minor groups from the major group as well as the capacity to select relevant features which enables the separation of minor groups although they are not representative in terms of quantity in the original dataset. In case of multi-class anomaly detection, the first step consists in evaluating the performance of the method to separate normal observations to anormal ones. The second step consists in evaluating the performance of the method to separate anormal observations between them.
\end{remark}

\begin{remark}
Note that only the evaluation part is divided in two steps. The choice of the features and clustering are done on the whole dataset, as illustrated in Figure \ref{schema_evaluation}. 

\begin{figure}[!h]
\caption{Scheme illustrating the process of 2 steps of the clustering evaluation}
\centering
\includegraphics[width=0.8\textwidth]{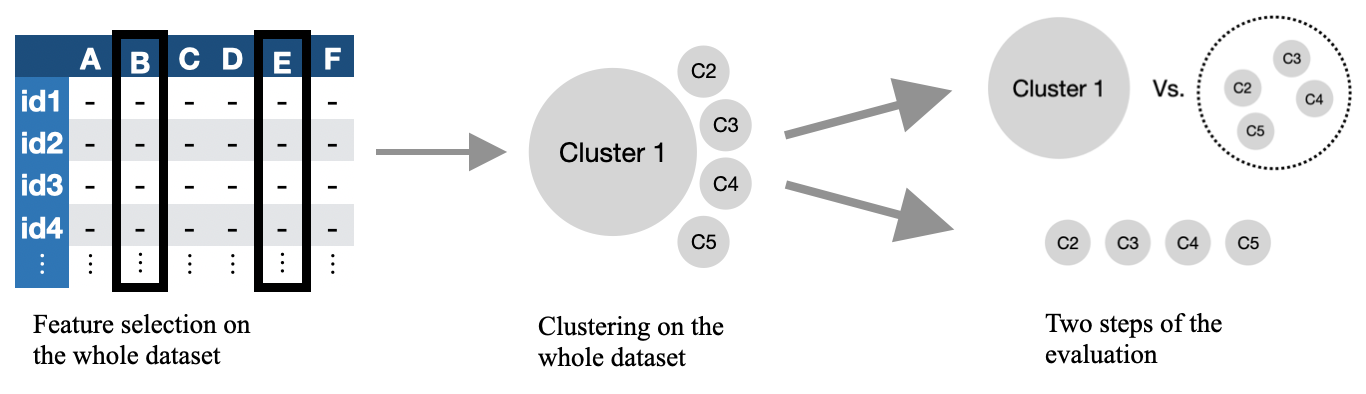}
\label{schema_evaluation}
\end{figure}

\end{remark} 

 \subsubsection{Clustering}\label{clustering}
The clustering task is done with the K-means algorithm because of its efficiency and popularity. The action is repeated 5 times and the results are averaged in order to minimise the different initialization effects. The standard deviation is also provided. Note that the clustering is done on all samples without train/test split since the objective is not to evaluate the performance of the clustering algorithm.

 \subsection{Number of features}\label{nfeatures}
Filter methods for feature selection assign a relevance score for each feature but do not provide a criteria to determine how many features we should retain. In general, it is advised to consider only few features to improve computational time and explainability. We propose to examine the evolution of the feature score to estimate the suitable number of features to select. \\

\begin{figure}[!h]
\caption{Evolution of score in Dataset 1}
\centering
\includegraphics[width=0.8\textwidth]{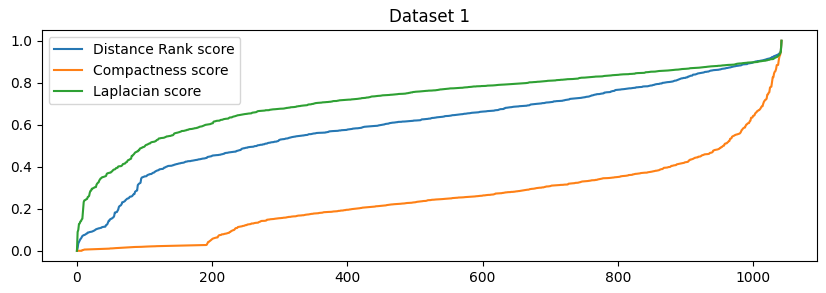}
\label{evol_scoreD1}
\end{figure}

\begin{figure}[!h]
\caption{Evolution of score in Dataset 2}
\centering
\includegraphics[width=0.8\textwidth]{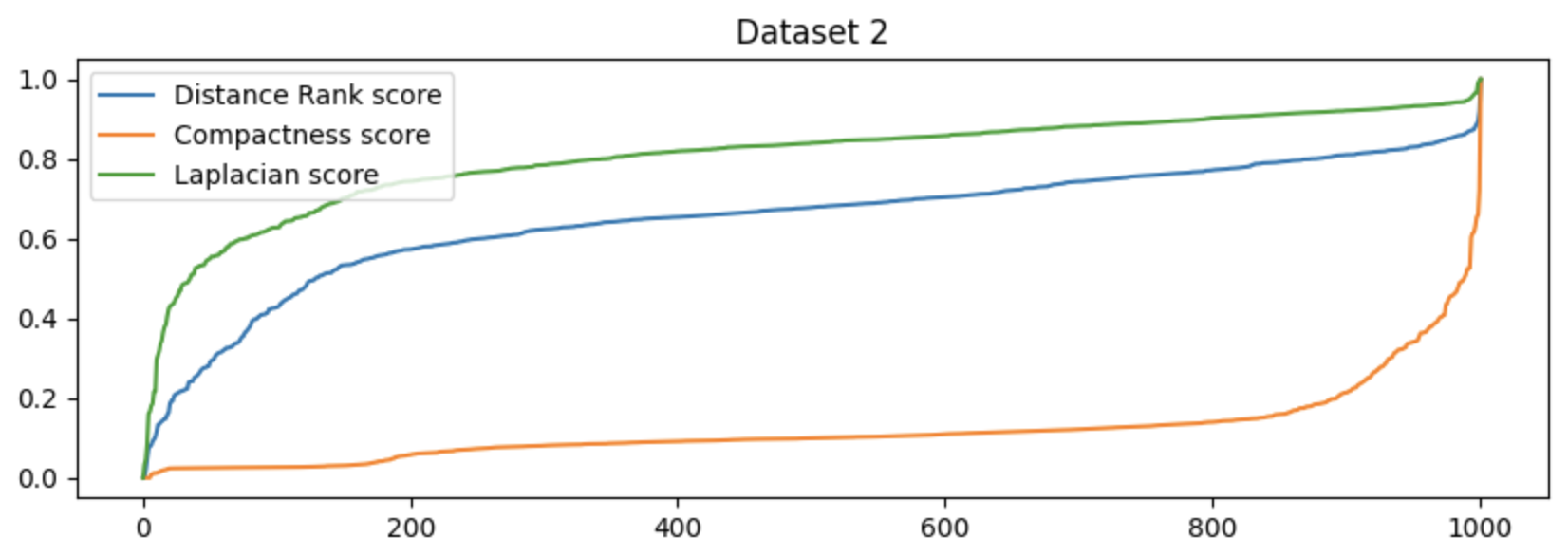}
\label{evol_scoreD2}
\end{figure}

\begin{figure}[!h]
\caption{Evolution of score in dataset MNIST5m. Same trend was observed for data MNIST5}
\centering
\includegraphics[width=0.8\textwidth]{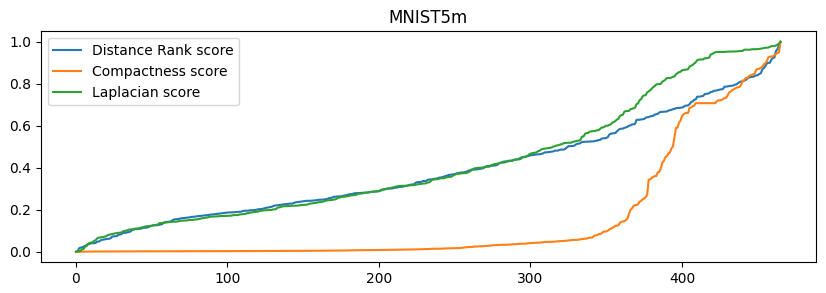}
\label{evol_scoreMNIST5m}
\end{figure}

\begin{figure}[!h]
\caption{Evolution of score in dataset orlraws10P}
\centering
\includegraphics[width=0.8\textwidth]{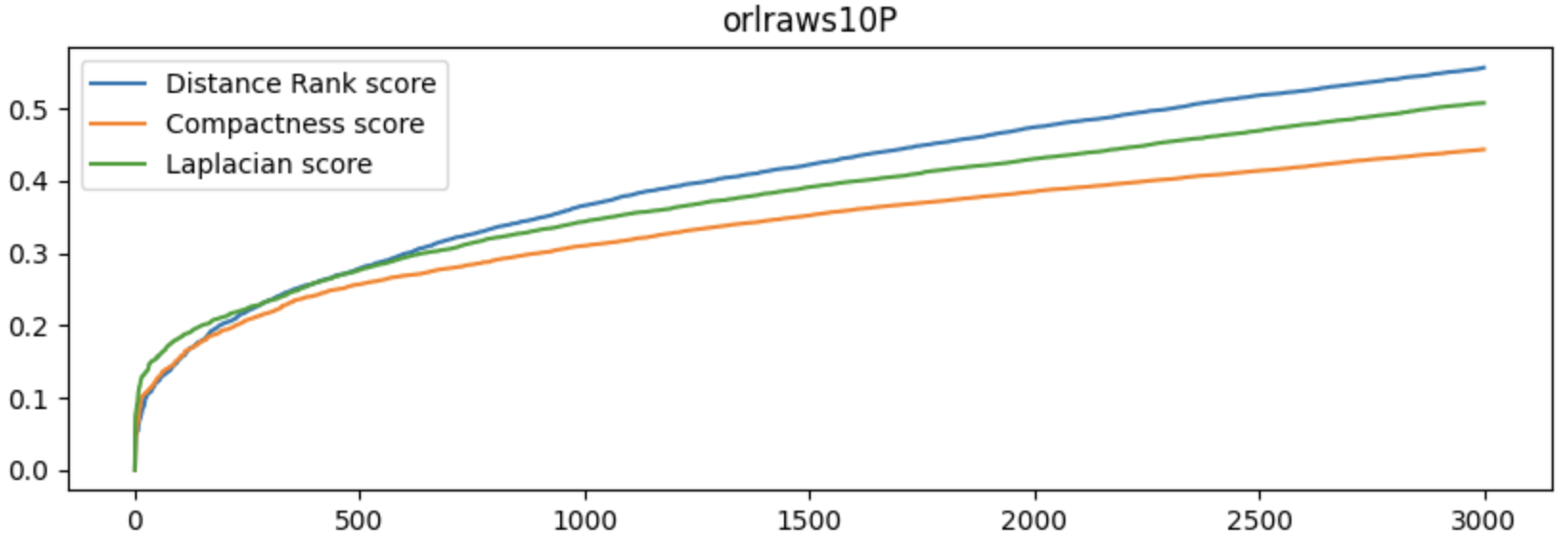}
\label{evol_scoreorlraws}
\end{figure}

Figures \ref{evol_scoreD1}, \ref{evol_scoreD2}, \ref{evol_scoreMNIST5m} and \ref{evol_scoreorlraws} display the score evolution of the ranked features for the four datasets. X-axis represents the feature rank in ascending order and y-axis represents the score for a given feature according to the tested method. Note that values are rescaled to the range from 0 to 1 and Distance Rank Score values are subtracted from 1 in order to be consistent with other methods where a low score indicates a high relevance. One way to choose the number of relevant features is to fix a threshold and to select features whose score is below this threshold. The concave shape of Distance Rank and Laplacian scores suggests that these methods would lead to select a few number of features, which is not the case for Compactness score.\\

Another option than setting a threshold consists in applying an heuristic elbow rule to estimate the suitable number of features to select. For example in the case of {\it Dataset 1}, {\it Dataset 2} and {\it orlraws10P} and according to the Distance Rank and Laplacian method, the ideal number of features to select is around 100 (see Figures \ref{evol_scoreD1}, \ref{evol_scoreD2} and \ref{evol_scoreorlraws}). 

 {\it Dataset 1} and {\it Dataset 2} are inspired by an explainability problem of anomaly detection \cite{explainable_ad}. In the following sections we select 20 and 10 features respectively, which seems to be a more reasonable quantity to be able to describe and characterize the various clusters in a comprehensive way. We also study the performance of the different methods when the number of selected features is higher, as suggested by the elbow rule.

\subsection{Results - imbalanced case}\label{results_imbalanced}

This subsection is devoted to imbalanced datasets. More precisely, we consider one large homogenous group which is completed by groups with few observations of different kind.  The feature selection algorithm needs to find the parameters characterizing the clusters of all types of observations, even those which are not highly represented.
Table \ref{imbalanced_total} compares the clustering performance of different methods on three imbalanced datasets by using the $F^W_1$-score (see equation \eqref{eq_FW}), showing that Distance Rank outperforms or is competitive with the other methods. \\

As mentioned in subsection \ref{clustering_evaluation}, imbalanced datasets also occur in the context of anomaly detection, i.e. when different types of misclassifications have not the same impact on the studied process. Since the total score does not provide detailed information on the origin of the errors, we display the results of two evaluation steps, as mentioned in subsection \ref{clustering_evaluation}.

Table \ref{imbalanced1st} compares the clustering performance of the methods according to the first step of evaluation in case of imbalanced datasets, which can be interpreted as the capacity to isolate minor groups from the major group. Distance Rank method is the best or is comparable with the other methods according to $F^W_1$ metric for the three considered datasets. It is especially performant on datasets with clearly distinguishable clusters, like it is a case on simulated {\it Dataset 1} and {\it Dataset 2}.

Table \ref{imbalanced2nd} compares the clustering performance of the methods according to the second step of evaluation in case of imbalanced datasets, which can be interpreted as the capacity to select the features characterizing each type of observation, even if some of these observations are not highly represented in the considered dataset. The Distance Rank method gives better results than the other methods when studying {\it Dataset 1} and {\it MNIST5m}. Since minor groups are of comparable size, we consider Accuracy and NMI metrics to evaluate the performance of the methods. \\

\begin{table}[!ht]
  \caption{Clustering performance on whole dataset when using different methods for feature selection.}
  \label{imbalanced_total}
\begin{center}
\begin{tabular}{|c|c|c|c|}
  \hline
   \multicolumn{4}{|c|}{ Performance - total }\\\hline\hline
 Data name & \multicolumn{1}{|c|}{Dataset 1} & \multicolumn{1}{|c|}{Dataset 2} & \multicolumn{1}{|c|}{MNIST5m} \\ 
  \hline
Eval. metric  & $F^W_1$ &  $F^W_1$ &  $F^W_1$  \\ 
 \hline
All features &  0.33 (0.01) 		 &  0.61 (0.01)				& 0.40 (0.01) \\ 
 \hline 
MaxVariance & 0.19 (0.02) 		 &  0.34 (0.04)				&  \textbf{0.41} (0.005) \\ 
 \hline
Laplacian score &  0.55 (0.004) 		 &  0.67 (0.001)				& 0.35 (0.02) \\ 
\hline
Compactness score &  0.22 (0.003) 		 &  0.37 (0.05)			& 0.37 (0.03) \\ 
\hline
Distance Rank score  &   \textbf{0.98} (0.003)	 	 &  \textbf{0.96} (0.0)				& 0.40 (0.02) \\ 
\hline
\end{tabular}
\end{center}
\end{table}

\begin{table}[!ht]
  \caption{Major group vs others - Clustering performance when using different methods for feature selection.}
  \label{imbalanced1st}
\begin{center}
\begin{tabular}{|c|c|c|c|}
  \hline
   \multicolumn{4}{|c|}{ Performance I}\\\hline\hline
 Data name & \multicolumn{1}{|c|}{Dataset 1} & \multicolumn{1}{|c|}{Dataset 2} & \multicolumn{1}{|c|}{MNIST5m} \\ 
  \hline
Eval. metric  & $F^W_1$ &  $F^W_1$ &  $F^W_1$  \\ 
 \hline
All features &  0.31 (0.02) 		 &  0.57 (0.01) 				& 0.42 (0.02)  \\ 
 \hline 
MaxVariance & 0.20 (0.02)	 	& 0.36 (0.04) 				&   \textbf{0.43} (0.005) \\ 
 \hline
Laplacian score &  0.50 (0.004) 		 & 0.62 (0.001)		& 0.37 (0.02)   \\ 
\hline
Compactness score &  0.23 (0.01) 		 & 0.39 (0.05) 		& 0.39 (0.03)  \\ 
\hline
Distance Rank score  &  \textbf{0.997} (0.003) 	 &  \textbf{0.98} (0.002)	 & 0.40 (0.02) \\ 
\hline
\end{tabular}
\end{center}
\end{table}

\begin{table}[!ht]
  \caption{Clustering performance within minor groups when using different methods for feature selection. }
    \label{imbalanced2nd}
\begin{center}
\resizebox{\textwidth}{!}{%
\begin{tabular}{|c|cc|cc|cc|}
  \hline
   \multicolumn{7}{|c|}{Performance II }\\\hline\hline
 Data name & \multicolumn{2}{|c|}{Dataset 1} & \multicolumn{2}{|c|}{Dataset 2} & \multicolumn{2}{|c|}{MNIST5m} \\ 
  \hline
Eval. metric  &  NMI & ACC &    NMI & ACC & NMI & ACC \\ 
 \hline
All features   & 0.78 (0.03)  & 0.56 (0.05)		 & 0.86 (0.05) & 0.84 (0.1)					 & 0.01 (0.02) & 0.26 (0.01) \\ 
 \hline 
MaxVariance  &  0.43 (0.04) &	0.33 (0.03)		 &    0.21 (0.002) & 0.43 (0.02)				&  0.01 (0.02) & 0.28 (0.01)\\ 
 \hline
Laplacian  score & 0.84 (0.001)  & 0.74	 (0.001)		  & 0.88 (0.001) & \textbf{0.89} (0.001)				  &  0.11 (0.03) & 0.30 (0.01) \\ 
\hline
Compactness score  &  0.49 (0.02) & 0.34 (0.05) 				&   0.21 (0.01) & 0.45 (0.01)				 & 0.05 (0.01) & 0.28 (0.01) \\ 
\hline
Distance Rank score  &   \textbf{0.90} (0.01) &\textbf{0.79} (0.02)			 & \textbf{0.90} (0.03) & 0.76	 (0)			 &  \textbf{0.23} (0.07) &  \textbf{0.43} (0.04) \\ 
\hline
\end{tabular}}
\end{center}
\end{table}

The computations in Tables \ref{imbalanced_total}, \ref{imbalanced1st} and \ref{imbalanced2nd} on {\it Dataset 1}, {\it Dataset 2} and {\it MNIST5m} have been done with 20, 10 and 30 features respectively. Mean and standard deviation of 5 executions are displayed.
Table \ref{increasing_n_features} compares, in the context of anomaly detection, how the performances of the methods change when the number of features increases. Distance Rank method is the most performant or competitive with other methods which suggest that this method is efficient to determine smaller subsets of relevant features.
It is an advantage to have a restricted number of features, especially in a situation when we want to describe the various clusters in a comprehensive way and to explain why a group of observations differs from the others.

\begin{table}[!ht]
  \caption{Clustering performance of the methods according to $F^W_1$ metric for increasing number of features. The two evaluation steps are denoted by I and II (see \ref{clustering_evaluation}).}
  \label{increasing_n_features}
\begin{center}
\resizebox{\textwidth}{!}{%
\begin{tabular}{|c|ccccccc|ccccccc|ccccccc|}
  \hline
 Data name  & \multicolumn{7}{|c|}{Dataset 1} 		& \multicolumn{7}{|c|}{Dataset 2} 	& \multicolumn{7}{|c|}{MNIST5m} \\ 
  \hline  
 N. of features  & 5 & 10 & 20 & 30 & 50 & 75 & 100  & 5 & 10 & 20 & 30 & 50 & 75 & 100  & 5 & 10 & 20 & 30 & 50 & 75 & 100 \\ 
 \hline
MaxVar I  &  0.2  & 0.19 &0.17 &0.2  &0.17& 0.2  &0.19		&  0.42 &0.32& 0.33 &0.32& 0.33 &0.32 &0.34 &	{\bf 0.44} &{\bf 0.43}& {\bf 0.4} & {\bf 0.43}& {\bf 0.45} &0.38 &0.37	 \\   
Laplacian I   &0.3  &0.47 &0.5  &0.6 & 0.92 &{\bf 0.82 } & {\bf 0.66 } & 		{\bf 0.78} &0.62 &0.66 &0.53& 0.53 &0.53 &0.53 &		0.39 &0.34& 0.38 &0.35 &0.33 &0.43& 0.4 	  \\ 
Distance Rank I    & {\bf 0.98} & {\bf 1.  } & {\bf 1. }  &{\bf 1. } &{\bf 1. } & 0.76 &0.64 &  		0.71 &{\bf 0.98} &{\bf 1. }&  {\bf 0.99}& {\bf 0.97}& {\bf 1.}  & {\bf 1.} &  		0.35 &0.42 &0.38& 0.4&  0.43 &{\bf 0.48} &{\bf 0.46 } \\
Compactness I   &	0.18 &0.18 &0.23& 0.22 &0.24 &0.25 &0.27 &	 	0.5  &0.36 &0.34& 0.33 &0.33 &0.32 &0.34&		 0.43 &0.37 &0.44 &0.39& 0.4&  0.38 &0.37 \\
 \hline  
MaxVar II  &  0.36 &0.36& 0.3 & 0.36 &0.33& 0.26& 0.26 			& 0.36 &0.37 &0.4 & 0.42& 0.43 &0.42 &0.33&		0.24 &0.3 & 0.25 &0.16& 0.17 &0.17& 0.2\\ 
 Laplacian II  & 0.21 &0.43 &0.67 &0.76 &{\bf 1. } &   {\bf 1. }  &  {\bf 1. }   & 			0.74 &{\bf 0.86} &{\bf 0.98} &{\bf 0.92}& 0.73 &0.77& 0.77 &  		0.15 &0.19 &0.19& 0.19& 0.15 &{\bf 0.19}& {\bf 0.19}   \\ 
Distance Rank II   & {\bf 0.65 } & {\bf 0.8 } & {\bf  0.79} &{\bf  0.79 }& 0.75 &0.75 &0.75&  		 {\bf  1. }  &0.78 &{\bf 0.98} &0.9 & {\bf 0.99} &{\bf 0.99 }&{\bf 0.99} 	&{\bf  0.25}& {\bf 0.29} &{\bf 0.32}& {\bf 0.36}& {\bf 0.29}& 0.14 &0.15   \\ 
 Compactness  II & 0.44 &0.36 &0.24 &0.27& 0.23& 0.23 &0.18&	        0.35 &0.42 &0.43& 0.47 &0.46& 0.39 &0.39 &		0.23 &0.2  &0.2  &0.16 &0.24 &0.15 &0.16 \\
  \hline 

\end{tabular}}
\end{center}
\end{table}

\subsection{Results - balanced case}\label{results_balanced}

This subsection deals with balanced datasets, i.e. we consider several types of observations while each type is represented equally. The objective is to reduce the number of original variables to continue the analyses with an acceptable number of variables. \\

\begin{remark}
We emphasize that studying balanced datasets is not equivalent to the second step of evaluation presented above. In case of imbalanced datasets the relevant features are selected once for all when considering all observations (major and minor groups). The case of balanced datasets would be equivalent to the second step of evaluation if new relevant features were selected when considering only abnormal observations. 
\end{remark} 

Table \ref{table_balanced} compares the clustering performance of the methods in the case of two balanced datasets. We notice that in this case the Distance Rank method is not the most powerful one, regarding either {\it MNIST5} or {\it orlraws10P}. Regarding {\it MNIST5}, we obtain the best results using all available features. It may suggest that the choice of the number of features is not suitable and has led to a loss of information.

\begin{table}[!ht] 
  \caption{Efficiency of clustering using different methods for feature selection. Mean and standard deviation of 5 executions are displayed. 100 and 30 features were used for {\it orlraws10P} and {\it MNIST5}, respectively. }
  \label{table_balanced}
\begin{center}
\begin{tabular}{|c|cc|cc|}
  \hline
   \multicolumn{5}{|c|}{Clustering performance - balanced clusters}\\\hline\hline
 Data name & \multicolumn{2}{|c|}{orlraws10P } & \multicolumn{2}{|c|}{MNIST5} \\ 
  \hline
Eval. metric  &   NMI & ACC & NMI & ACC \\ 
 \hline
All features &   0.79 (0.03) &  0.71 (0.05)	 &   \textbf{0.70} (0.002) & \textbf{0.87} (0.001)  \\ 
 \hline 
MaxVariance & \textbf{0.81} (0.02) &  0.72 (0.04) & 	  0.64 (0.001) & 0.48 (0.0003) \\ 
 \hline
Laplacian score &   0.79 (0.01) &   \textbf{0.73} (0.03)		 & 0.49 (0.001) & 0.65 (0.001)  \\ 
\hline
Compactness score & 0.80 (0.03) &  \textbf{0.73} (0.04)&   				0.44 (0.002) & 0.62 (0.001) \\ 
\hline
Distance Rank score &   0.68 (0.03) &  0.59 (0.05)			&  0.3 (0.001) & 0.49 (0.004)  \\ 
\hline
\end{tabular}
\end{center}
\end{table}

\subsection{Optimisation}\label{optimisation}
Laplacian and Compactness methods consider the local neighbors of each observation to compute the feature scores. The search of the nearest neighbors is optimised and thus these methods gain an advantage in terms of computational time in comparison with methods such as Distance Rank which consider all observations for the computations. 
The complexity of the Distance Rank method grows linearly with number of dimensions but grows in a quadratic way in the number of observations.\\

In order to reduce the computational time of our method, we examine the efficiency of the feature selection when considering data subsets. Ideally, if the observations chosen to subsample are selected randomly and if the subsample is not too small, each type of observation should be represented, hence the algorithms should have enough information to determine the relevant features.\\

Table \ref{subset} shows the results of clustering when the features are selected using Distance Rank and Laplacian score when using 50 \% and 30 \% of the observations of the original datasets. To present concise results, only Laplacian score, which is the most competitive one of the tested methods, has been kept. For simplicity, we display the results of the first evaluation step in case of imbalanced classes measured with $F^W_1$ metric. Other results have shown the same trend.
We notice that the efficiency of our feature selection method on subset depends on the dataset, but does not decrease significantly with the subsampling.
Further research on optimisation of the computation time without efficiency loss is needed to make the new method more competitive.

\begin{table}[!ht]
  \caption{Comparison of the computational time and clustering performance when the process of feature selection is done on a subset. The $F^W_1$ score on subsamples is averaged on 5 executions, each execution using a different subsample for feature selection while the performance is measured on whole dataset. Time is measured in seconds as an average of 5 executions of feature selection algorithm on a subset of corresponding size. As in the previous section, 20, 10 and 30 features have been selected for {\it Dataset 1, Dataset 2} and {\it MNIST5m}, respectively. }
  \label{subset}
\begin{center}
\resizebox{\textwidth}{!}{%
\begin{tabular}{|c|cc|cc|cc|cc|cc|cc|}
  \hline
 Data name  & \multicolumn{4}{|c|}{Dataset 1}  & \multicolumn{4}{|c|}{Dataset 2} & \multicolumn{4}{|c|}{MNIST5m} \\ 
  \hline  
  Method  & \multicolumn{2}{|c|}{Distance Rank} & \multicolumn{2}{|c|}{Laplacian}  & \multicolumn{2}{|c|}{Distance Rank} & \multicolumn{2}{|c|}{Laplacian} & \multicolumn{2}{|c|}{Distance Rank} & \multicolumn{2}{|c|}{Laplacian}  \\ 
    \hline  
 Eval. metric  & $F^W_1$  & Time & $F^W_1$  & Time  & $F^W_1$  & Time & $F^W_1$  & Time & $F^W_1$  & Time & $F^W_1$  & Time \\ 
 \hline  \hline
100 \% 	&  {\bf  0.99} & 17.6  &	  0.50 & {\bf  0.046}  	 	& {\bf 0.98} & 21.70 	& 0.61  &  {\bf 0.045}		& {\bf 0.41} & 3.2 		& 0.37 & {\bf 0.022}  	  \\ 
 \hline
50 \%  	&   {\bf 0.99} & 4.25   &   0.46 & {\bf  0.015}	& {\bf 0.99} &   4.69  	&  0.60 &  {\bf 0.016} 	& {\bf 0.43} &   0.84    	 &  0.40 & {\bf 0.008}   \\ 
 \hline
30 \%  	&  {\bf 0.93} &	1.61   & 0.55 & {\bf 0.008} 	& {\bf 0.94} & 1.73  	&  0.45 & {\bf 0.009} 		&0.42 & 0.350   	&   {\bf 0.44} & {\bf 0.006}  \\ 
 \hline
\end{tabular}}
\end{center}
\end{table}

\section{Conclusion}\label{conclusion}

This paper has presented a filter method for unsupervised feature selection based on Spearman's Rank Correlation between distances on the observations and on feature values. Our method is particularly efficient in case of imbalanced datasets. Moreover, the explainability of the method is reinforced compared to other methods since only a few relevant features are needed to achieve a good performance. More research on the automatic determination of the suitable number of features and optimisation of the computational time is needed. Improvements of the performance in case of balanced datasets is also an objective of further research.

\bibliographystyle{alpha}
\bibliography{Bib_articleFS}

\end{document}